\documentclass[11pt]{article}

\usepackage{graphicx}

\usepackage{times}
\usepackage{listings}  
\usepackage{amsmath}

\begin{document}

\lstset{
numbers=left, 
numberstyle=\small, 
numbersep=8pt, 
frame = single, 
framexleftmargin=15pt}

\title{Walk a Mile in Their Shoes: a New Fairness Criterion for Machine
Learning}

\author{
   Norman Matloff \\
   matloff@cs.ucdavis.edu \\
   Department of Computer Science\\
   University of California, Davis\\
   Davis, CA 95616, USA
}

\maketitle

\begin{abstract}  

\noindent 
The old empathetic adage, ``Walk a mile in their shoes,'' asks that one
imagine the difficulties others may face. This suggests a new ML
counterfactual fairness criterion, based on a \textit{group} level: How
would members of a nonprotected group fare if their group were subject
to conditions in some protected group?  Instead of asking what sentence
would a particular Caucasian convict receive if he were Black, take
that notion to entire groups; e.g. how would the average sentence for
all White convicts change if they were Black, but with their same White
characteristics, e.g.\ same number of prior convictions?  We frame the
problem and study it empirically, for different datasets.  Our approach
also is a solution to the problem of covariate correlation with
sensitive attributes.  

\end{abstract} 

\section{Introduction}
\label{intro}

The issue of fairness in machine learning (ML) is the subject of
increasing concern, with many interesting algorithms proposed, both for
fair prediction and for assessing fairness.  As always, though, ``The
devil is in the details.''  Exactly what do we mean by ``fair''?
Many criteria for fairness have been proposed.  See for instance
\cite{barocasbook} \cite{vermaexplained} \cite{berk} \cite{barocaslum}
\cite{goel} for overviews, some of them written from a critical point of
view.

Much work has been done on fair ML from counterfactual points of view.
What sentence would this Caucasian convict receive if he had been Black
but otherwise with the same relevant characteristics?  See \cite{loftus}
for a comprehensive review.  In this paper, a novel approach
to the counterfactual is proposed.

\subsection{Relation to the Legal Realm}

Having served as an expert witness in a number of discrimination cases in
litigation (e.g.\ \cite{reid}), I view an important point in choosing
fairness criteria to be consideration of legal issues, including legal
standards for evidence but even more important, development of methodology
that is easily understood by judges and juries.

\cite{xiang}, \cite{sunstein} and \cite{bao} give detailed analyses of
how fairness criteria developed in the ML literature may or may not be
consistent with US federal statutes and case law.  The legal situation
in the European Union is apparently less precisely formulated at this
point \cite{wachter}, but will certainly evolve in the near future;
this may produce constraints.  See \cite{india} and \cite{shin} for
interesting South and East Asian perspectives.

A key point regarding statistical analyses presented in litigation is
the Daubert standard \cite{florida}) \cite{gastwirth} \cite{kaye}.  The
vast majority of fair ML methods do not account for sampling
variability, i.e.\ do not include mechanisms for significance testing
and confidence intervals.\footnote{The authors in \cite{hurlin}
speculate that they are the first to have developed such mechanisms; a
literature search seems to confirm this.} Accordingly, most published
methods may be vulnerable to challenge in court.

Anohter key point in litigation is that the statistical analysis presented be
intuitively clear and reasonable to judges and juries (\cite{competent}).  
Indeed, in the case of judges, the US Federal Judicial Center, a federal
agency chaired by the Chief Justice of the Supreme Court, found the
issue of statistical literacy pressing enough to commission a guide to
statistical methods for judges \cite{guide}.  This was viewed as
essential even though presumably most judges had some exposure to the
subject in school; that will not be the case for some, likely most,
jurors.  It is thus imperative that fair ML methodology be easily
grasped on an intuitive level, a major goal of this paper.

There is also concern among many regarding \textit{intersectionality},
the pernicious interaction between the separate discriminatory actions
against different protected groups \cite{foulds}, \cite{fredman2}.
As explained in \cite{foulds},

\begin{quote}

Intersectionality is a lens for examining societal unfairness which
originally arose from the observation that sexism and racism have
intertwined effects, in that the harm done to Black women by these two
phenomena is more than the sum of the parts

\end{quote}

Methodology aimed at fairness must take intersectionality into account.
Linear models of effects of race, gender and aga, for instance, may
require inclusion of interaction terms.  

\subsection{Contributions of This Paper}

Although much of the presentation here will be motivated by legal
aspects, the need for fair ML goes far beyond the legal realm.  A firm
or government agency, for instance, may be concerned that its AI tool
may be effective at addressing some objective function, but at the same
time may be unfair to some important groups.  

Indeed, no single fairness criterion is best for all possible
applications.  Thus it is desirable for ML analysts to have available a
variety of fairness criteria to choose from.  

This paper proposes a new fairness criterion, motivated
specifically by these desiderata:

\begin{itemize}

\item [(a)] A criterion should be easily relatable to not only ML
analysts, but also to non-STEM academics, the press, participants in
litigation, and the general public.

\item [(b)] As such, a criterion should be kept simple, describable easily in
plain language.  

\item [(c)] Conditional distributions can be confusing.  Fairness
criteria with lesser explicit involvement with conditional quantities
may be preferred.

\item [(d)] Again in the interest of simplicity, a criterion should not
consist of many numbers; it should consist of a single number for each
protected (base group, counterfactual group) pair.

\item [(e)] A criterion should include a mechanism for generating associated
standard errors, thus enabling statistical inference.

\item [(f)] A criterion should be easily adaptable to intersectionality
analyses.

\item [(g)] A criterion should properly account for covariates that may be
related both to the outcome of intereset and to status as a member of a
protected group.

\end{itemize}  

Concerning that last point, let $X$ and $S$ denote our feature vector
and indicator of protected status, respectively.  Let $Y$ be the outcome
of interest, and let $\widehat{Y}$ denote the
predicted value of $Y$, given $X$.  Ideally we wish that $\widehat{Y}$
is independent of $S$, or nearly so.  This goal is obstructed by the
fact that there may be features in $X$ that are correlated with $S$, so
that the model makes some use of $S$ after all.  Let's call this the
\textit{X,S correlation problem}.  

\subsection{Organization of This Paper}

As seen in the paper's title and abstract, the proposed method is
inspired by the old empathetic adage, ``Walk a mile in their shoes,''
which asks that one imagine the difficulties others may face.
Accordingly the proposed fairness criterion will be called Walk a Mile
(WaM).

WaM will be motivated in Section \ref{wam}, then described in detail in
Section \ref{method}.  In Section \ref{empir}, we illustrate the
criterion on various real datasets.  Section \ref{var} will be devoted
to mathematical derivations and empirical methods to assess the
statistical accuracy of WaM.  Extensions of WaM, such as for
false-positive rates, will be discussed in Section \ref{extend}.
Finally, Section \ref{discuss} consists of conclusions and discussion of
future work.

\section{Introducing WaM}
\label{wam}

WaM takes the counterfactual view of a protected group \textit{as a
whole}.  As a quick example, consider some data from the 2000 US census
(similar to the Adult dataset, another common example in fair ML
studies).  WaM would yield statements like,

\begin{quote}

Hey, you men out there!  Your mean income in this study was \$63554.04.
But if you were women of the same occupation, age and education, your
mean would be only \$53829.78.

\end{quote}

Or, in COMPAS, an algorithm for assisting judges in sentencing of
defendants:

%
%
%
%
%
%

\begin{quote}

Take note, Caucasian defendants!  Your mean decile risk score is 3.67.
(Smaller values are better.) But if you were Black while having the
\textit{same characteristics} as yours, your mean would be 4.19.  

\end{quote}

This directly addresses the X,S correlation problem, in a novel way.
Unlike methods that attempt to remove the influence of $S$ in $X$, we
\textit{adjust} for that influence.

In the employment example above, in which $S$ is gender, the occupation
variable in $X$ is strongly related to $S$.  This variable codes for one
of six computer-professional occupations in the study, and is strongly
gendered:


\begin{tabular}{rrrrrrr}
  \hline
 & A & B & C & D & E & F \\ 
  \hline
men & 0.20 & 0.22 & 0.34 & 0.02 & 0.04 & 0.17 \\ 
women & 0.31 & 0.23 & 0.33 & 0.04 & 0.03 & 0.06 \\ 
   \hline
\end{tabular}

Men are only 2/3 as likely as women to be in occupation A, while the
male rate for occupation F is nearly triple that of women.

And of course there is considerable variation in pay across occupations
in the study:

\begin{tabular}{rrrrrrr}
  \hline
 & A & B & C & D & E & F \\ 
  \hline
mean \$ & 50396 & 51374 & 53640 & 67019 & 68798 & 69494 \\ 
   \hline
\end{tabular}

WaM adjusts for this by equalizing the distribution of $X$:  How would
men fare with \textit{their} distribution of $X$ if they were women?

\section{Why WaM?}

Other than the desideratum regarding standard errors, most of the
multitude of existing fairness criteria satisfy at least several of the
desiderata listed above.  Yet, not all the desiderata are important in
all application settings.  With that in mind, let us discuss some of the
desiderata, 

\subsection{Use of Conditional Quantities}

Desideratum (c) states,

\begin{quote}

Conditional distributions can be confusing.  To the degree possible, 
fairness criteria that do not explicitly involve conditional quantities
are preferred.

\end{quote}

This one may be surprising, as a number of fairness criteria are based on
explicit conditional quantites, such as the true positive rate
(TPR, conditional probability) and regression functions (conditional mean)
\cite{barocasbook}.  

Many such criteria are used successfully in practice, such as in 
\cite{kozodoi}.  There a credit-equity measure is proposed,

\begin{equation}
SP = 
\frac{1}{2} 
|
(FPR_{x_{\alpha}=1} - FPR_{x_{\alpha}=0}) + 
(FNR_{x_{\alpha}=1} - FNR_{x_{\alpha}=0})
|
\end{equation} 

This may be effective in a corporate internal introspective document,
but difficult to comprehend by most jurors,

It is well known that conditional quantities are subject to frequent
misunderstanding \cite{prodromou}.  Such confusion is not limited to
classroom settings, and indeed arises often in ordinary discourse.  The
authors in \cite{kramer} cite a number of misleading newspaper
headlines, as well as court caess.  After the Covid-19 pandemic emerged,
for instance, it was common to see news article with titles such as
``Coronavirus Testing:  What Is a False Positive rate?'' (\cite{bbc}),
and there was much confusion over the efficacy of policy
(\cite{morris}).  

Careful presentation, say in describing data to a jury, may succeed
in ameliorating such problems, but it is desirable to have available
criteria that less explicitly cite conditional quantities.  Though WaM
technically does involve conditional quantities, it does so in an
implicit manner that is easier for non-STEM people to grasp.

It should be noted in passing that if a fairness analysis involves
ranks, other major problems arise \cite{lahoti}, as two individuals
may have roughly equal traits yet be far apart in rankings. 

\subsection{Dimensionality of the Fairness Criterion}

Desideratum (d) says,

\begin{quote}

Again in the interest of simplicity, a criterion should not
consist of many numbers; it should consist of a single number for each
protected (base group, counterfactual group) pair.

\end{quote}

Many methods involve setting up multiple scenarios of real or
hypothetical individual cases, with a typical scenario being say,
``Consder a 36-year-old single woman applying for credit, with a current
bank balance of \$1200, a yearly income of \$63,250'' and so on.  Again,
for an internal, introspective corporate analysis, this finely detailed
approach may be informative, indeed necessary.  But in many settings,
again such as litigation, it is best to keep to a single summary number.
WaM satisfies this crtierion that our fairness measure consists
of a single number, as opposed to measures involving pointwise
counterfactual fairness, such as \cite{loftus}.  

Computing a single summary number is especially useful in situations in
which multiple policies, multiple ML prediction methods and so on are
being compared over multiple protected groups.  If possible, having just
a single number for each case in the comparison is desirable.

On the other hand, however, note Einstein's famous quote,
``Everything should be made as simple as possible, but not simpler.''
Some settings are fundamentally complex, and it will be seen below that
WaM can help uncover more complex fairness problems.

\subsection{Ability of the Consumer (e.g.\ Juror) to Empathize}

It is well known in legal circles that empathy is crucial in judges and
jurors.  As pointed out in \cite{linder}, this is especially important
when, for instance, jurors and a defendant are of a different race.
Presumably this holds for gender as well.

The proposed method's name, Walk a Mile, is motivated by empathetic
aims, arising from the saying, ``Walk a mile in their shoes.''
\cite{mueller} recounts the history of this phrase, which goes back at
least as early as the Cherokee tribe of Native Americans.

WaM is aimed at capturing this spirit, helping the intended recipient of
the presentation, say a jury, relate emotionally.  The hypothetical
example given earlier in this paper,

\begin{quote}

Hey, you men out there!  Your mean income in this study was \$63554.04.
But if you were women of the same occupation, age and education, your
mean would be only \$53829.78.

\end{quote}

should be something that strikes a chord with either men or women.
Men, for instance, may be somewhat detached when presented with
evidence of discrimination against women, but may better relate to
comparisons such as the above, which bring in their own gender on a
groupwide level.

\subsection{Amenability to Intersectionality Issues} 

%
%
%

The analyst may find that accounting for intersectional issues may
complicate implementation for some types of fairness criteria.  But
intersectionality flows naturally from the definition of WaM, with no
special extensions or special cases required.  An example will be
presented shortly.

\subsection{Further Comments}

It is not claimed here that WaM uniquely satisfies the desiderata
(again, ignoring the standard errors issue).  Yet, different application
settings may have different goals, and the novel method presented here
should be very effective in a number of settings.  The WaM approach to
measuring fairness should have simple, immediate relevance in many
discussion venues, such as policymaking, litigation, social science
research and so on.  It thus should be part of any analyst's toolkit.

\section{Notation and Background}

As above, let $X$ be a vector of features, such as credit history and
jobs, and let $S$ indicate membership in a sensitive group, such as a
racial minority.  

Let $Y$ denote the score value of interest---say a credit rating or a
sentence given to a convicted criminal.  Or, the ``score'' may be a
direct outcome, such as income in our example above.  Note that though
one could distinguish between an outcome $Y$ and a decision $D$ as in
\cite{barocaslum}, this is not done here, in the interests of
simplicity; indeed, for convenience let us refer to $Y$ as a ``policy,''
whether a formal one such as the decile risk score, or a virtual one
imposed by society such as via discriminatory wages.  For binary $Y$, we
assume coding as a dummy/one-hot variable, coded 0,1.

Note that while $S$ will typically be a scalar, such as gender or race,
it could be an intersectional vector, consisting of gender \textit{and}
race.  In the vector case, assume for notational convenience that $S$
has been recoded to scalar form, in which $S$ is categorical, with $s$
values.  If $S$ is continuous, such as age, assume that it has been
discretized to $s$ values.

Though WaM may remind some of \textit{covariate shift} methods, such as
\cite{polo}, \cite{zhang}, its goals and operation are actually opposite
to those methods.  They are concerned with \textit{adapting} a fitted ML
model in the face of future distributional shifts in data to be
predicted.  In WaM, we \textit{utilize} the fact that the distribution
of $X | S$ varies with $S$, as a means of measuring unfairness..

One more point to get started:  Let $U$ and $V$ be any random variables
for which the quantities below exist.  Here and below we will be using
the Law of Total Expectation,

\begin{equation}
EV = E[ E(V|U) ]
\end{equation}

and the Law of Total Variance (``Pythagorean Theorem" for variance)
\cite{ross},

\begin{equation}
\label{totvar}
Var(V) = E[Var(V|U)] + Var[E(V|U)]
\end{equation}

\section{Details of the Method}
\label{method}

Now, in the context of the probabilistic generative process of our
data, let $X_i,~ i=1,...,s$ denote a random vector of length $p$
having the distribution of $X | S = i$.  Then define

\begin{equation}
\mu_i(t) = E(Y | X_i = t), ~ i=1,...,s
\end{equation}

Here are the quantities that WaM estimates,

\begin{equation}
\nu_{ij} = E[\mu_j(X_i)],~ i,j=1,...,s
\end{equation}

Then $\nu_{ij}$ is the mean outcome that members of group $i$
would have if they were subjected to the policies used for group $j$.

In the case $i = j$, we simply have

\begin{equation}
\label{samegrp}
\nu_{ii} =  E[\mu_i(X_i)] 
= E[(E(Y|X) ~|~ S_i]
= E(Y | S =i)
\end{equation}

using the Tower Theorem for conditional expectations (\cite{klenke}).
That is, $\nu_{ii}$ is the mean (non-counterfactual) outcome in group $i$. 

These quantities $\nu_{ij}$ must be estimated from our data.  Let
$X_{ik}$ denote the $k^{th}$ realization of $X_i$ in group $i$, $k =
1,...,n_{i}$, and let $Y_{ik}$ denote the associated $Y$ value.  In
other words, considering only the rows $R_i$ of our training data for which
$S = i$, then $X_{ik}$ is the value of $X$ in the $k^{th}$ such row,
and $Y_{ik}$ is the value of $Y$ in that row.  Let $N_i$ denote the
number of such rows, with $n$ rows in all.

We fit one's favorite predictive algorithm, say linear regression or
random forests, taking the training set to be $R_i$, yielding
the estimated regression function for group $i$,
$\widehat{\mu}_i(t)$.  We can then form the counterfactual outcomes at
the individual level,

\begin{equation}
\widehat{Y}_{jk} = \widehat{\mu}_i(X_{jk})
\end{equation}

Our WaM estimates are then

\begin{equation}
\widehat{\nu}_{ij} = \sum_{k=1}^{N_i} \widehat{\mu}_j(X_{ik})
\end{equation}

For the purpose of forming $\widehat{\mu}_{i}(t)$., linear or logistic
regression, k-nearest neighbors, gradient boosting,and random forests
all work in a straightforward manner.  Neural networks are also fine for
continuous $Y$; in the case of categorical $Y$, the final-stage
activation function must be chosen appropriately to produce class
probabilities.

SVM also is fine for the categorical $Y$ case, provided calibration
methods are used on the output scores \cite{huang}.  For continuous $Y$,
various forms of support vector regression are available, even for
advanced forms such as quantile regression \cite{svr}. 

It is important to note that unlike many ML fairness criteria, WaM has
no problem with unbalanced data, since $\widehat{\mu}_{i}(t)$ is
estimated separately for each $i$.  However, one does of course need the
generative process underlying $(X_{ik},Y_{ik})$ to be that of
$E(Y | X_i)$.

Other than of course requiring that the expectations exist, there are
really no statistical assumptions underlying WaM.

\section{Empirical Evaluation}
\label{empir}

Here we illustrate WaM via various datasets, using both
parametric regression methods, either linear or logistic, and random
forests or k-Nearest Neigbhors.  The software is available at
\textit{https://github.com/matloff/WAMfair}.

The examples are presented solely for illustrating the workings of the
method.  No feature engineering is performed, and default values are
used for hyperparameters.  

\subsection{Census Data}

This is a dataset from the 2000 US Census, focused on Silicon Valley.
Variables are age, education, occupation (among six programmer and
engineer job categories), gender, wage income, and number of weeks worked.

First, a look at gender (example from above), with a linear model:

\begin{table}[ht]
\centering
\begin{tabular}{rrr}
  \hline
 & men cf.. & women cf.\\ 
  \hline
men, act. & 63554.04 & 53829.78 \\ 
  women, act. & 58646.87 & 50403.48 \\ 
   \hline
\end{tabular}
\end{table}

We use ``act.'' and ``cf.'' to denote the actual and counterfactual
cases.  Women as a group had a mean wage of about \$50,403, but if they
had been men with the same characteristics, their mean would have been
\$58,647.

The random forest analysis is similar:

\begin{table}[ht]
\centering
\begin{tabular}{rrr}
  \hline
 & men cf. & women cf. \\ 
  \hline
men act. & 63554.04 & 53728.65 \\ 
  women act. & 59464.30 & 50403.48 \\ 
   \hline
\end{tabular}
\end{table}

(The diagonal elements do not change in this second analysis, due to
(\ref{samegrp}).)

\subsection{Employment Race Test Data} 
\label{lak}

This data is from \cite{bertrand}.  Re'sume's were sent to random
employers, with the same qualifications but some with ``Black-sounding''
given names, such as Lakisha.  $Y$ is an indicator variable recording
whether the employer called the applicant after receiving the re'sume'.
A number of variables were used, such as education, years of
experience, military background and so on.


Only the variables 
"education", "ofjobs", "yearsexp", "honors",
"volunteer", "military", "empholes", "workinschool" ,
"email", "computerskills", "race", "call" and
"adid" were used.

Here is the random forests analysis:

\begin{table}[h]
\centering
\begin{tabular}{rrr}
  \hline
act & bcf & wcf \\ 
  \hline
b & 0.06 & 0.05 \\ 
w &   0.03 & 0.10 \\ 
   \hline
\end{tabular}
\end{table}

We see that ``Caucasian'' applicants had a 10\% callback rate, but would have
had only a 3\% rate if they had been Black with the same qualifications.
Interestingly, though ``Black'' applicants did not seem to benefit from
having ``White'' names.

As noted, WaM automatically includes intersectional cases, no extension
or special cases involved.  Again with random forests: 

\begin{table}[ht]
\centering
\begin{tabular}{rrrrr}
  \hline
 & fbcf & mbcf & fwcf & mwcf \\ 
  \hline
fbact & 0.07 & 0.06 & 0.05 & 0.10 \\ 
  mbact & 0.03 & 0.10 & 0.05 & 0.11 \\ 
  fwact & 0.02 & 0.05 & 0.06 & 0.04 \\ 
  mwact & 0.02 & 0.05 & 0.03 & 0.09 \\ 
   \hline
\end{tabular}
\end{table}

Here 'fb' means female Black, 'mb' means male Black, and so on.  So,
Black women would get a boost from being White men, with their actual
7\% callback rate increasing to 10\%, but not, for instance, from being
Black men.  The latter, on the other hand, would benefit by being White
men, but not much by being White women.

The latter result, of course, may depend on the job type, and a more
segmented analysis is needed, together with domain expertise, but WaM
does seem to have uncovered some interesting, complex relationships.



\subsection{COMPAS}

%

We used the variables 'age', 'juv\_fel\_count', 
'decile\_score', 'juv\_misd\_count',  'juv\_other\_count', 
'priors\_count', 'sex' and 'race'.  Due to insufficient data, 
the Asian and Native American instances were changed to Other.

Using a linear regression model to predict decile score, we have

\begin{tabular}{rrrrr}
  \hline
 & African-American.cf & Caucasian.cf & Hispanic.cf & Other.cf \\
  \hline
African-American.act & 5.28 & 4.78 & 4.41 & 4.13 \\
  Caucasian.act & 4.19 & 3.67 & 3.20 & 2.94 \\
  Hispanic.act & 4.36 & 3.77 & 3.32 & 3.07 \\
  Other.act & 4.36 & 3.76 & 3.31 & 3.05 \\
   \hline
\end{tabular}

We showed two of these numbers in Section \ref{intro} above, with
possible implications that the COMPAS system rates Black defendants
unfairly highly.  Various other numbers in the table suggest that as
well.  

The random forest analysis is similar:

\begin{tabular}{rrrrr}
  \hline
 & African-American.cf & Caucasian.cf & Hispanic.cf & Other.cf \\
  \hline
African-American.act & 5.28 & 4.73 & 4.28 & 4.16 \\
  Caucasian.act & 4.15 & 3.67 & 3.20 & 3.00 \\
  Hispanic.act & 4.27 & 3.74 & 3.32 & 3.07 \\
  Other.act & 4.26 & 3.70 & 3.22 & 3.05 \\
   \hline
\end{tabular}

What about gender?

\begin{table}[ht]
\centering
\begin{tabular}{rrr}
  \hline
 & Female & Male \\ 
  \hline
Female & 4.05 & 3.60 \\ 
  Male & 5.04 & 4.50 \\ 
   \hline
\end{tabular}
\end{table}


The COMPAS algorithm seems to have a substantial bias against women.
Their mean decile score was 4.05, but would have been only 3.60 had they
been men with the same characteristics.

%

\subsection{German Credit Data}

This is another commonly used example in ML fairness research, though
one of the aspects considered below is less common.  Let's start with
the standard aspect, investigating possible gender bias.  Using a
logistic model for a `Good' credit rating, we have

\begin{tabular}{rrr}
  \hline
 & female.cf & male.cf \\ 
  \hline
female.act & 0.72 & 0.64 \\ 
  male.act & 0.70 & 0.65 \\ 
   \hline
\end{tabular}

Women in general had a 72\% rate of being assigned a Good rating, but if
they were men with the same characteristics, only 64\% would get this
score.  There may be bias against men here.  The results using random
forests are similar:

\begin{tabular}{rrr}
  \hline
 & female & male \\ 
  \hline
female & 0.72 & 0.66 \\ 
  male & 0.72 & 0.65 \\ 
   \hline
\end{tabular}

Let's take a look at (discretized) age, using random forests:

\begin{tabular}{rrrr}
  \hline
 & (18.9,37.7].cf & (37.7,56.3].cf & (56.3,75.1].cf \\ 
  \hline
(18.9,37.7].act & 0.68 & 0.75 & 0.70 \\ 
  (37.7,56.3].act & 0.74 & 0.75 & 0.71 \\ 
  (56.3,75.1].act & 0.68 & 0.74 & 0.72 \\ 
   \hline
\end{tabular}

Possibly a slight bias in favor of the middle-aged consumers, less
favorable to the young and old.

Is there discrimination against foreign workers in the credit realm?
Again, running an analysis with random forests, we have

\begin{tabular}{rrr}
  \hline
 & not for.cf & for.cf \\ 
  \hline
not for.act & 0.89 & 0.74 \\ 
  for.act & 0.84 & 0.69 \\ 
   \hline
\end{tabular}

About 69\% of foreign workers have Good credit, but if they were
citizens with the same characteristics, 84\% would be judged good credit
risks.  So, there may indeed be some discrimination here.  

%

\section{Assessment of Sampling Variance}
\label{var}

Claims of discrimination are sensitive.  As such, there must be some
indication of the accuracy of the fairness criterion used, in the form
of statistical standard errors (``error bars'').  This is vital in
academia, the corporate world and especially in litigation, where,
as previously noted, case law requires it.

Let's recall the notation for our data:  Define $Y$, $X$ and $X_i$ as
before.  Denote the $k^{th}$ data point in group $i$ by $X_{ik}$ and
$Y_{ik}$, respectively.  Data points are i.i.d.\ across $k$ for fixed
$i$, and independent across $i$.  Let $n_i$ denote the number of data
points in group $i$.  Vectors are written as columns.

\subsection{Exact Expression for Variance in the Linear Case}

For notational convenience, it is assumed here that all data has been
centered, thus having mean 0.  This eliminates the intercept term for
the coefficent vector, and the need to have a 1s column in the design
matrix.

Here we have

\begin{equation}
E(Y | X_i) = 
\beta_i^T {X}_i
\end{equation}

where $\beta_i = 
(\beta_{i1}, 
\beta_{i2},..., 
\beta_{ip})^T
$
and T denotes matrix transpose.

In this setting,

\begin{equation}
\nu_{ij} = E(\beta_j^T {X}_i) = \beta_j^T E({X}_i)
\end{equation}

Using OLS, we obtain an unbiased linear estimate $\widehat{\beta_j}$ of
$\beta_j$.  Denote its covariance matrix by $Cov(\widehat{\beta_j})$
(computed via the standard formula and available for instance in R via
the \lstinline{vcov()} function).  We are not assuming normal $Y_{ij}$,
but under our i.i.d.\ conditions $\widehat{\beta_j}$ will be
asymptotically normal.

Then our estimated $\nu_{ij}$ is

\begin{equation}
\label{estnuij}
\widehat{\nu}_{ij} = \widehat{\beta}_j^{~T} {A}_i
\end{equation}

where our sample estimate of $E{X}_i$ is

\begin{equation}
{A}_i = \frac{1}{n_i} \sum_{k=1}^{n_i} {X}_{ik}
\end{equation}

Then conditioning on $A_i$ and using (\ref{totvar}), we have

\begin{eqnarray}
\label{big1}
Var \left (\widehat{\beta_j}^T {A}_i  \right ) &=&
Var \left ( A_i^T \widehat{\beta}_j \right ) \\
&=& E \left [ Var( A_i^T \widehat{\beta}_j | A_i ) \right  ] +
   Var \left [ E( A_i^T \widehat{\beta}_j) | A_i \right  ] \\
&=& E \left [
   A_i^T Cov(\widehat{\beta}_j) A_i]
   \right ] +
   Var \left [
   A_i^T \beta_j
   \right ] \\
&=& E \left [
   A_i^T Cov(\widehat{\beta}_j) A_i]
   \right ] +
   \beta_j^T Cov(A_j) \beta_j \\
&=& E \left [
A_i^T Cov(\widehat{\beta}_j) A_i]
\right ] +
\frac{1}{n_i}
\beta_j^T Cov(X_i) \beta_j 
\label{whew}
\end{eqnarray}

where $Cov(A_i)$ and $Cov(X_i)$ are the covariance matrices of
$A_i$ and $X_i$, respectively.  Hence we have the exact expression for
variance of WaM.

These quantities can then easily be estimated directly from their sample
analogs.  In fact, our estimate for (\ref{whew}) is simply

\begin{equation}
{A}_i^T \widehat{Cov}(\widehat{\beta}_j) A_i] +
\frac{1}{n_i}
\widehat{\beta}_j^T \widehat{Cov}(X_i) \widehat{\beta}_j 
\label{stderr}
\end{equation}

where $\widehat{Cov}$ denotes covariance matrices estimated from the
data.

Note that in (\ref{estnuij}) we have the product of a quantity
$\widehat{\beta}_i$ that converges in distribution (to a multivariate
Gaussian distribution with mean $\beta_{i}$) and a quantity $A_j$ that
converges in probability (to $EX_i$).  In other words,
$\widehat{\nu}_{ij}$ too is asymptotically normal (\cite{klenke}), with
standard error given by the square root of (\ref{stderr}), and
confidence intervals can be formed.

\subsection{The Bootstrap}

For nonparametric models, say random forests, one may apply bootstrap
methods (\cite{wager}).  One does resampling of the data many time,
generating many values of $\widehat{\nu}_{ij}$.  Our new point estimate
is the average of these values, and the standard error is their standard
deviation.\footnote{The software computes only bootstrap results, not
the formula derived above for the linear case.}

\subsection{Examples}

We earlier broke down the COMPAS data by gender, finding an apparent
bias against women.  Here are the standard errors, based on 100
resamplings:

\begin{table}[ht]
\centering
\begin{tabular}{llrrrr}
  \hline
act & cf & myGrpEst & theirGrpEst & bias & biasSE \\ 
  \hline
Female & Male & 4.11 & 3.65 & 0.46 & 0.05 \\ 
  Male & Female & 4.60 & 5.16 & -0.56 & 0.08 \\ 
   \hline
\end{tabular}
\end{table}

So for instance an approximate 95\% confidence interval for that 0.46
figure would be $0.46 \pm 1.96 \times 0.05$, or about 0.36 to 0.56.
The bias is real.

As another example, consider the Employment Race Test Data, Section
\ref{lak}.  Again with 100 resamples, the bootstrap results were

\begin{table}[ht]
\centering
\begin{tabular}{llrrrrrr}
  \hline
act & cf & myGrpEst & myGrpSE & theirGrpEst & theirGrpSE & bias & biasSE \\ 
  \hline
b & w & 0.06 & 0.01 & 0.06 & 0.01 & 0.00 & 0.01 \\ 
  w & b & 0.10 & 0.01 & 0.04 & 0.00 & 0.06 & 0.01 \\ 
   \hline
\end{tabular}
\end{table}

So there is a genuine difference in callback ratings.  If the White
applicants were to apply under ``Black'' names with the same
employment-relevant characteristics, their chances of being called back
by employers would only be about 40\% as high.

\section{Extensions}
\label{extend}

The WaM approach can be extended, applying it to various common 
fairness criteria.  For instance, we could address questions such as:

\begin{quote}
Consider loan applications, predicting default, with the criterion false
positive rate.  If Caucasian applicants were to have their present
qualifications but were Black, what would their false positive rate be?
\end{quote}

Assume and the classification rule (if distributions were fully known)

\begin{equation}
\widehat{Y} =
\begin{cases}
1, & \mu_j(X_i) \geq 0.5 \\
0, & \mu_j(X_i) < 0.5 \\
\end{cases}
\end{equation}


Let's use $\gamma_{ij}$ to denote the false positive rate that those in
group $i$ would experience if they were subjected to the treatment of
those in group $j$.  The quantities being estimated are

\begin{equation}
\gamma_{ij} = 
\frac
   {E \pmb{\{} 1_{[0.5,1]} \pmb{[} \mu_j(X_i) ~\cdot~ (1-\mu_j(X_i)) \pmb{]} 
      \pmb{\}} }
   {E \left [ 1_{[0.5,1]} [\mu_j(X_i)] \right ]}
\end{equation}

These are then estimated by their sample analogs.

For instance, consider a Dutch census dataset (\cite{dutch}).  The
outcome of interest is whether a respondent's occupation is considered
prestigious, using various demographic, economic and household
characteristics.  Here are the results using a k-Nearest Neighbors
analysis: 

\begin{table}[ht]
\centering
\begin{tabular}{rllrrrr}
  \hline
 & male c.f. & female c.f.& myGrpEst & theirGrpEst & bias & biasSE \\ 
  \hline
male & 1 & 2 & 0.19 & 0.22 & -0.03 & 0.01 \\ 
  female & 2 & 1 & 0.21 & 0.22 & -0.01 & 0.01 \\ 
   \hline
\end{tabular}
\end{table}

For example, the FP rate for men was about 19\%, but would be 22\% if
they were women with the same characteristics.

\section{Accounting for Model Edge Bias}

Fit of one's regression model is important for WaM.  Standard fit
assessment techniques may be used, but here we discuss some issues
concerning bias at the edges of one's dataset.

To motivate this discussion, think of a simple setting in which we use
k-Nearest Neighbors to predict human weight from height.  Consider a
data point in which, say, the person is unusually tall.  Then his/her
neighbors are likely to be shorter than this person, thus lighter.  In
other words, the prediction on this data point may be biased downward.

The situation with random forests is similar.  The leaf in a decision tree 
into which this hypothetical tall person falls is also a neighborhood,
and again, the other points in this leaf will likely be shorter/lighter.

But remedies have been developed for both k-NN \cite{knnedgebias} and
random forests \cite{grf}, in the form of calculating a linear fit
within a neighborhood or leaf.  These will soon be incorporated in our
WaM software.

\section{Conclusions and Future Work}
\label{discuss}

We have introduced a new approach to fairness in machine learning, a
group-counterfactual measure we call WaM.  It serves as a simple measure
of fairness that can be easily grasped by nontechnical people, and it is
a solution to the X,S correlation problem.

We believe WaM opens the door to extensive further research.  We mention
two possible areas:

\begin{itemize}

\item Extension to other fairness and predictive power measures:  In
Section \ref{extend}, we extended WaM to False Positive Rates.  There
are various other rates that may be of interest in WaM's
group-counterfactual approach.  AUC may also be extended to WaM.

\item As explained earlier, any report of a fairness measure in ML
should be accompanied by a standard error.  This is especially important
in litigation, but also vital in any context in which the report will be
used in decision making.

The bootstrap solution we presented here is convenient, but is time
consuming.  Moreover, it cannot easily be used in pre-planning, e.g.\ of
sample size, needed in a study.  For such purposes, explicit formulas
for estimator variance such as (\ref{whew}) can be helpful, and worthy
of further study.  Extension of (\ref{whew}) to the generalized linear
model, notably the logistic and Poisson regression cases, should be
pursued. 

\end{itemize}

\bibliographystyle{acm}
\bibliography{Paper}

\begin{thebibliography}{10}

\bibitem{grf}
{\sc Athey, S., Tibshirani, J., and Wager, S.}
\newblock {Generalized random forests}.
\newblock {\em The Annals of Statistics 47}, 2 (2019), 1148 -- 1178.

\bibitem{bao}
{\sc Bao, M., Zhou, A., Zottola, S., Brubach, B., Desmarais, S., Horowitz, A.,
  Lum, K., and Venkatasubramanian, S.}
\newblock It's compaslicated: The messy relationship between rai datasets and
  algorithmic fairness benchmarks, 2021.

\bibitem{barocasbook}
{\sc Barocas, S., Hardt, M., and Narayanan, A.}
\newblock {\em Fairness and Machine Learning: Limitations and Opportunities}.
\newblock 2021.

\bibitem{berk}
{\sc Berk, R., Heidari, H., Jahbari, S., Kearns, M., and Roth, A.}
\newblock Fairness in criminal justice risk assessments: The state of the art.
\newblock {\em Sociological Methods and Research 50\/} (2018).

\bibitem{bertrand}
{\sc Bertrand, M., and Mullainathan, S.}
\newblock Are emily and greg more employable than lakisha and jamal? a field
  experiment on labor market discrimination.
\newblock {\em American Economic Review 94}, 4 (September 2004), 991--1013.

\bibitem{goel}
{\sc Corbett-Davies, S., and Goel, S.}
\newblock The measure and mismeasure of fairness: A critical review of fair
  machine learning, 2018.

\bibitem{reid}
{\sc Court~of Appeal, Sixth~District, C.}
\newblock Reid v. google inc, 2007.

\bibitem{knnedgebias}
{\sc Elizabeth~Yancey, R., Xin, B., and Matloff, N.}
\newblock Modernizing k-nearest neighbors.
\newblock {\em Stat 10}, 1 (2021), e335.
\newblock e335 sta4.335.

\bibitem{foulds}
{\sc Foulds, J.~R., and Pan, S.}
\newblock An intersectional definition of fairness.
\newblock {\em CoRR abs/1807.08362\/} (2018).

\bibitem{fredman2}
{\sc Fredman, S.}
\newblock Intersectional discrimination in eu gender equality and
  non-discrimination.

\bibitem{guide}
{\sc Freedman, D., and Kaye, D.}
\newblock Reference guide on statistics.
\newblock In {\em Reference Manual on Scientific Evidence}. 2011.

\bibitem{gastwirth}
{\sc Gastwirth, J.}
\newblock {\em Statistical Science in the Courtroom}.
\newblock Statistics for Social and Behavioral Sciences. Springer New York,
  2012.

\bibitem{huang}
{\sc Huang, Y., Li, W., Macheret, F., Gabriel, R.~A., and Ohno-Machado, L.}
\newblock A tutorial on calibration measurements and calibration models for
  clinical prediction models.
\newblock {\em Journal of the American Medical Informatics Association 27}, 4
  (2020), 621--633.

\bibitem{hurlin}
{\sc Hurlin, C., Perignon, C., and Saurin, S.}
\newblock The fairness of credit scoring models.

\bibitem{svr}
{\sc Hwang, C., and Shim, J.}
\newblock A simple quantile regression via support vector machine.
\newblock In {\em Proceedings of the First International Conference on Advances
  in Natural Computation - Volume Part I\/} (Berlin, Heidelberg, 2005),
  ICNC'05, Springer-Verlag, p.~512–520.

\bibitem{kaye}
{\sc Kaye, D.}
\newblock The dynamics of daubert: Methodology, conclusions, and fit in
  statistical and econometric studies.
\newblock {\em Virginia Law Review 87\/} (2001).

\bibitem{sunstein}
{\sc Kleinberg, J., Ludwig, J., Mullainathan, S., and Sunstein, C.~R.}
\newblock {Discrimination in the Age of Algorithms}.
\newblock {\em Journal of Legal Analysis 10\/} (04 2019), 113--174.

\bibitem{klenke}
{\sc Klenke, A.}
\newblock {\em Probability Theory: A Comprehensive Course}.
\newblock World Publishing Corporation, 2012.

\bibitem{kozodoi}
{\sc Kozodoi, N., Jacob, J., and Lessmann, S.}
\newblock Fairness in credit scoring: Assessment, implementation and profit
  implications.
\newblock {\em European Journal of Operational Research 297}, 3 (mar 2022),
  1083--1094.

\bibitem{kramer}
{\sc Kramer, W., and Gigerenzer, G.}
\newblock How to confuse with statistics or: The use and misuse of conditional
  probabilities.

\bibitem{loftus}
{\sc Kusner, M.~J., Loftus, J.~R., Russell, C., and Silva, R.}
\newblock Counterfactual fairness, 2018.

\bibitem{lahoti}
{\sc Lahoti, P., Gummadi, K., and Weikum, G.}
\newblock ifair: Learning individually fair data representations for
  algorithmic decision making.

\bibitem{linder}
{\sc Linder, D.}
\newblock Juror empathy and race.

\bibitem{florida}
{\sc Mahle, S.}
\newblock The impact of daubert v. merrell dow pharmaceuticals, inc., on expert
  testimony: with applications to securities litigation.
\newblock {\em Florida Bar Journal 73}, 3 (1999).

\bibitem{bbc}
{\sc Maybin, S., and Casserly, J.}
\newblock Coronavirus testing: What is a false positive?, 2020.

\bibitem{barocaslum}
{\sc Mitchell, S., Potash, E., Barocas, S., D'Amour, A., and Lum, K.}
\newblock Algorithmic fairness: Choices, assumptions, and definitions.
\newblock {\em Annual Review of Statistics and Its Application 8}, 1 (2021),
  141--163.

\bibitem{morris}
{\sc Morris, J.}
\newblock Israeli data: How can efficacy vs. severe disease be strong when 60%
  of hospitalized are vaccinated?, 2021.

\bibitem{mueller}
{\sc Mueller, S.}
\newblock Developing empathy: Walk a mile in someone’s shoes, 2020.

\bibitem{polo}
{\sc Polo, F.~M., and Vicente, R.}
\newblock Effective sample size, dimensionality, and generalization in
  covariate shift adaptation, 2021.

\bibitem{prodromou}
{\sc Prodromou, T.}
\newblock Secondary school students' reasoning about conditional probability,
  samples, and sampling procedures.

\bibitem{ross}
{\sc Ross, S.}
\newblock {\em A First Course in Probability}.
\newblock Pearson Prentice Hall, 2010.

\bibitem{india}
{\sc Sambasivan, N., Arnesen, E., Hutchinson, B., and Prabhakaran, V.}
\newblock Non-portability of algorithmic fairness in india, 2020.

\bibitem{shin}
{\sc Shin, D.~D.}
\newblock Toward fair, accountable, and transparent algorithms: Case studies on
  algorithm initiatives in korea and china.
\newblock {\em Javnost - The Public 26}, 3 (2019), 274--290.

\bibitem{competent}
{\sc Thompson, W.}
\newblock Are juries competent to evaluate statistical evidence?
\newblock {\em Law and Contemporary Problems 52\/} (1989).

\bibitem{vermaexplained}
{\sc Verma, S., and Rubin, J.}
\newblock Fairness definitions explained.
\newblock In {\em 2018 IEEE/ACM International Workshop on Software Fairness
  (FairWare)\/} (2018), pp.~1--7.

\bibitem{wachter}
{\sc Wachter, S., Mittelstadt, B., and Russell, C.}
\newblock Bias preservation in machine learning: The legality of fairness
  metrics under eu non-discrimination law.
\newblock {\em West Virginia Law Review 123}, 3 (2021).

\bibitem{wager}
{\sc Wager, S., Hastie, T., and Efron, B.}
\newblock Confidence intervals for random forests: The jackknife and the
  infinitesimal jackknife, 2014.

\bibitem{dutch}
{\sc {World Bank}}.
\newblock 'the dutch virtual census of 2001, ipums subset', 2001.

\bibitem{xiang}
{\sc Xiang, A., and Raji, I.~D.}
\newblock On the legal compatibility of fairness definitions, 2019.

\bibitem{zhang}
{\sc Zhang, T., Yamane, I., Lu, N., and Sugiyama, M.}
\newblock A one-step approach to covariate shift adaptation, 2021.

\end{thebibliography}

\end{document}